\documentclass{article}

    \usepackage[final]{neurips_2020}

\usepackage[utf8]{inputenc} 
\usepackage[T1]{fontenc}    
\usepackage{hyperref}       
\usepackage{url}            
\usepackage{booktabs}       
\usepackage{amsfonts}       
\usepackage{nicefrac}       
\usepackage{microtype}      
\usepackage[pdftex]{graphicx}

\title{Detecting Hateful Memes Using a Multimodal Deep Ensemble}

\author{%
  Vlad~Sandulescu \\
  Wunderman Thompson \\
  Copenhagen, Denmark \\
  \texttt{vlad.sandulescu@wundermanthompson.com} \\
}

\begin{document}

\maketitle

\begin{abstract}
  While significant progress has been made using machine learning algorithms to detect hate speech, important technical challenges still remain to be solved in order to bring their performance closer to human accuracy. We investigate several of the most recent visual-linguistic Transformer architectures and propose improvements to increase their performance for this task. The proposed model outperforms the baselines by a large margin and ranks 5\textsuperscript{th} on the leaderboard out of 3,100+ participants. 
  \footnote{Code is available at \url{https://github.com/vladsandulescu/hatefulmemes}.}
\end{abstract}

\section{Introduction}

The internet is having a huge impact on all of our lives and our virtual presence reflects both our personalities and beliefs but also our biases and prejudices. Billions of people are interacting with various online content every day and while some of it is highly useful and enriches our knowledge and understanding of the world, an increasing portion of this content is also harmful. This includes hate speech, misinformation and other forms of online abuse. An increasing amount of effort is required to quickly detect this content, scale up the review work and make automatic decisions to take down the harmful media fast in order to minimize the inflicted harm to the readers.

Many of our interactions happen on social media platforms, which we use to share messages and pictures with our private community or general public audiences. 
Facebook AI has launched a competition
\footnote{\url{https://www.drivendata.org/competitions/64/hateful-memes/}} to flag hateful memes consisting of both images and text. For this purpose they provide a unique labeled dataset of 10,000+ high quality new multimodal memes. The goal of the challenge is to create an algorithm that identifies multimodal hate speech in internet memes, while also being robust to their benign flip. A meme might be mean or hateful either because of the meme image itself, or the text or their combination. Benign flipping is an augmentation technique used by the competition organizers to flip a meme from hateful to non-hateful and viceversa. This requires changing either the meme text or the image to flip its label. Figure~\ref{fig-benign} shows how this process works.

Since the problem is formulated as a binary classification task, the primary evaluation metric used to rank the results is the area under the receiver operating characteristic curve (AUROC). This represents the area under the ROC curve, which in turn plots the True Positive Rate (TPR) vs. False Positive Rate (FPR) at different classification thresholds T. The goal is to maximize the AUROC. 
$$
\mathrm{AUROC}=\int_{\infty}^{-\infty} \mathrm{TPR}(T) \mathrm{FPR}^{\prime}(T) d T
$$
Accuracy is the secondary tracked metric and it calculates the percentage of instances where the predicted class \^{y} matches the actual class, y in the test set. 
$$
\mathrm{Accuracy}=\frac{1}{N} \sum_{i=0}^{N} I\left(y_{i}=\hat{y}_{i}\right)
$$
Ideally, the model maximizes both these metrics.

In summary, our contribution is threefold:
\begin{itemize}
  \item We experiment with and fine-tune both single stream state-of-the-art architectures: VL-BERT \citep{su2020vlbert}, VLP \citep{zhou2019unified}, UNITER \citep{chen2020uniter} and dual stream models: LXMERT \citep{tan2019lxmert} and compare their performance to the provided baselines. These transformers architectures are chosen such that we can leverage they were pre-trained on a wide spectrum of datasets.
  \item We propose a novel bidirectional cross-attention mechanism to couple inferred caption information with the meme caption text. This leads to increased classifier performance in detecting hateful memes. This approach is similar to the cross-attention between paired images used in \citep{chen2020uniter}.
  \item We show deep ensembles \citep{lakshminarayanan2017simple} improve on single model predictions by a significant margin.
\end{itemize}

\begin{center}
  \begin{figure}
    \centering
    \includegraphics[width=1\linewidth]{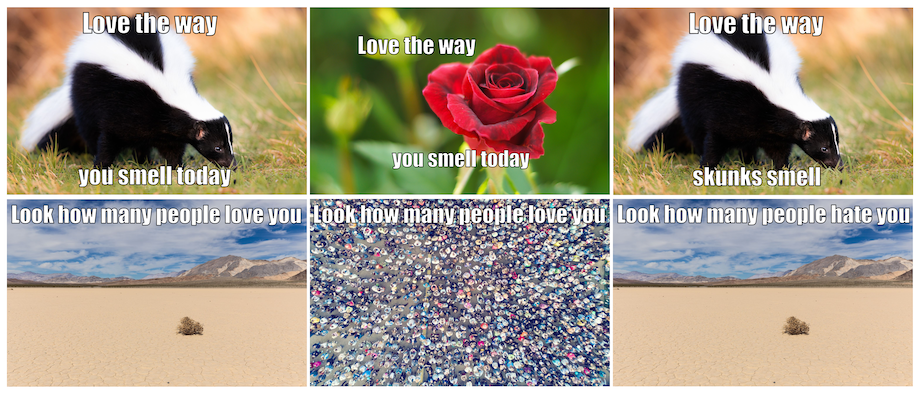}
    \caption{Multimodal memes and benign confounders. Hateful memes (left), benign image confounders (middle) and benign text confounders (right).}
  \end{figure}
\end{center}
\label{fig-benign}

\section{Related Work}

Transformer architectures pre-trained on large datasets have achieved state-of-the-art on numerous language processing problems. BERT \citep{devlin2019bert} is perhaps the most popular of these architectures because of its simplicity and strong performance. Training such large architectures on joint visual-linguistic embeddings has recently showed very promising results for visual-linguistic tasks such as visual question answering, visual reasoning and understanding and image captioning.
LXMERT \citep{tan2019lxmert} employed two single-modal networks (also known as a dual-stream architecture) to process text inputs and images, learning cross-modality encoder representations using a Transformer to combine the two information streams. The image features are extracted using a Faster R-CNN feature extractor \citep{Anderson2017up-down}. This is also used by the more recent single-stream architectures, VL-BERT \citep{su2020vlbert} and UNITER \citep{chen2020uniter}, which place a single Transformer on top of joint images-text embeddings. \citep{zhou2019unified} propose the first unified model for both visual understanding (e.g. visual question answering) and vision-language generation (e.g. image captioning). 
\citep{gomez2019exploring} built a large dataset for multimodal hate speech detection harvested from Twitter using specific hateful seed keywords. However they found the multimodal models they experimented with did not outperform the text unimodal ones.

\section {Method}

\paragraph{Leverage the diversity of the pre-training datasets.}
One of the goals of this research is to leverage the fact the Transformer single-stream and dual-stream models have been pre-trained on large datasets from a wide range of domains. The Transformer attention architectures are very successful in NLP tasks and the masked language modeling pre-training technique in BERT is both powerful and flexible. Empirical results show  the pre-training procedure can better align the visual-linguistic embeddings and benefit the downstream tasks, as mentioned before, such as visual question answering and visual commonsense reasoning. If pre-training a visual-linguistic Transformer architecture helps when fine-tuning for downstream tasks, perhaps ensembling multiple models pre-trained on different datasets might bring even more benefits?

Table~\ref{tab-pretraining-datasets} shows the pre-training datasets for each of the models.

\begin{table}
  \caption{Pre-training datasets for each model}
  \label{tab-pretraining-datasets}
  \centering
  \begin{tabular}{lllllllll}
    \toprule
         & Books corpus & CC & COCO & VG & SBU & GQA & VQA 2.0 & VG-QA \\
         & Wikipedia & & & & & & &   \\
    \midrule
    VL-BERT & X & X & & & & & &     \\
    VLP & & X & & & & & &     \\
    UNITER & & X & X & X & X & & &     \\
    LXMERT & & & X & X & & X & X & X    \\
    \bottomrule
  \end{tabular}
\end{table}

\begin{figure}
  \centering
  \includegraphics[width=1\linewidth]{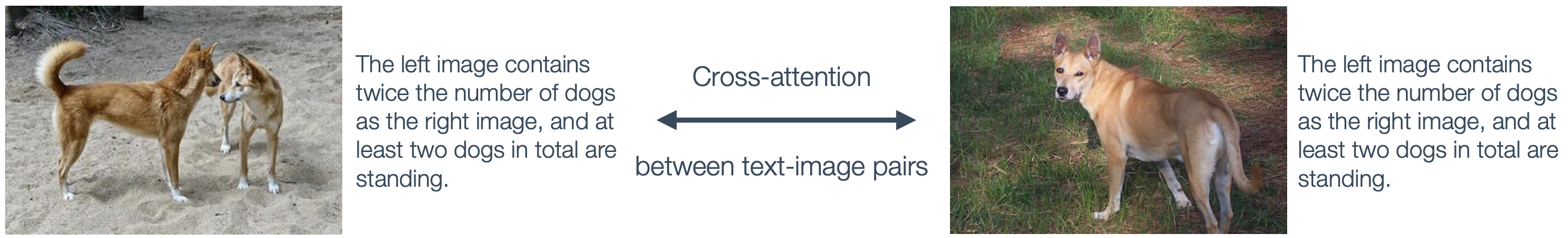}
  \caption{Sample from the NLVR\textsuperscript{2} dataset. Each caption is paired with two images.}
  \label{fig-nlvr}
  \includegraphics[width=1\linewidth]{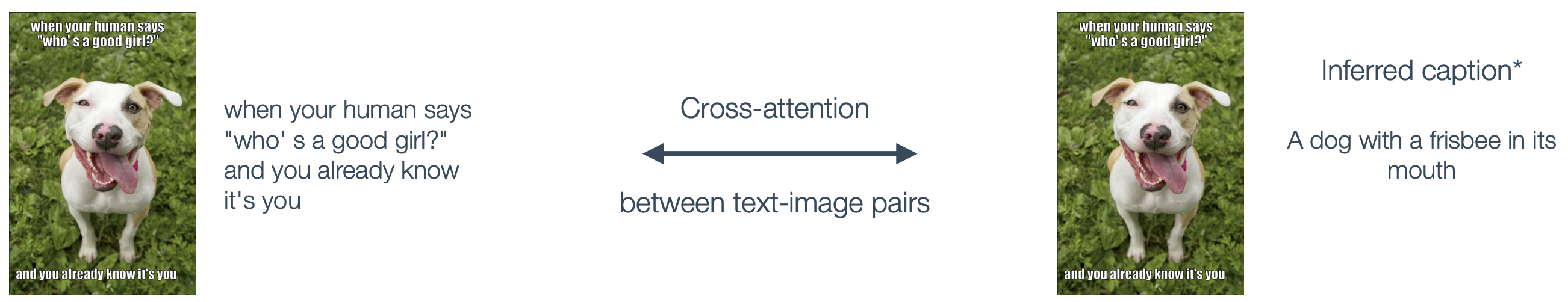}
  \caption{Training sample for the Hateful Memes dataset. Each meme image is paired with two texts.}
  \label{fig-inferred}
\end{figure}

\paragraph{UNITER with meme text plus inferred caption cross-attention.}
The Natural Language for Visual Reasoning for Real (NLVR\textsuperscript{2}) is an academic dataset containing human-written English sentences grounded in pairs of photographs \citep{suhr2019corpus}. The dataset consists of pairs of visually complex images coupled with a statement about the image pair and a binary label. UNITER placed amongst the top models in this challenge by adding a cross-attention module between text-image pairs, splitting each sample in two and repeating the text. Then they apply attention pooling to each of these sequence, concatenate them and finally add the classification head, a multi-layer perceptron. Similar to this approach, we propose instead to repeat the meme image in each half-sequence and add an inferred meme caption as the second text. We generate image captions using the Show and Tell model from \citep{Vinyals_2017}
\footnote{An implementation for doing image captioning inference using the pre-trained \citep{Vinyals_2017} model can be found at \url{github.com/HughKu/Im2txt}.}.
This way the model could learn not only from the meme text already provided, but also from the new captions generated from a model trained on a different corpus. Figures~\ref{fig-nlvr} and ~\ref{fig-inferred} show the cross-attention between image and text pairs.

\section {Experiments}

\begin{figure}
  \centering
  \includegraphics[width=1\linewidth]{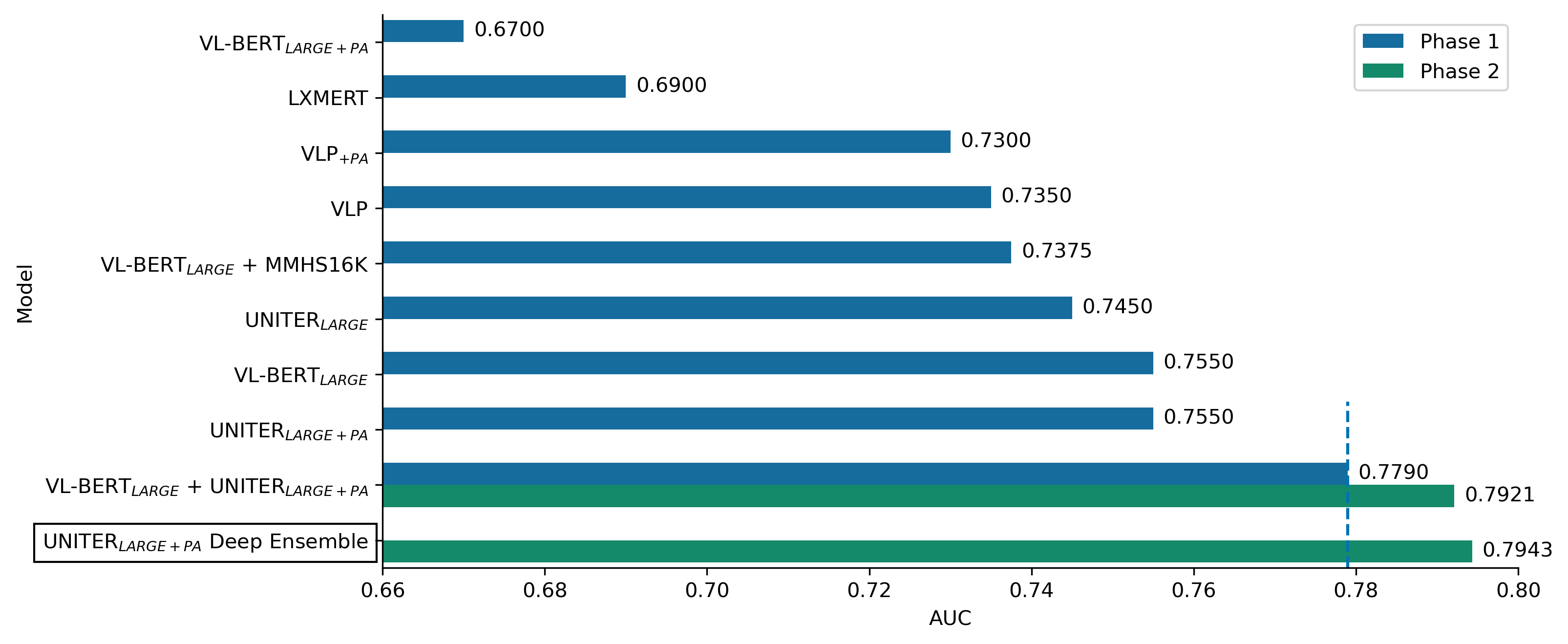}
  \caption{AUROC results for each model for both phases of the competition.}
  \label{fig-results}
\end{figure}

We run extensive experiments using all four pre-trained models: LXMERT, VLP, VL-BERT and UNITER. We adapt the bidirectional cross-attention using inferred captions for UNITER, VL-BERT and VLP. We do not use it with LXMERT, since its performance on the Hateful Memes dataset is low. Figure~\ref{fig-results} concisely summarizes our key experiments. 

We also experiment with the MMHS150K dataset from \citep{gomez2019exploring}. We heavily filter it as well as balance it, reducing it to only 16K samples, by avoiding to include cartoons memes for example or memes with too little text. We fine-tune VL-BERT\textsubscript{LARGE} using the MMHS16K dataset for four rounds, then fine-tune it further using the Hateful Memes dataset for another four. The results were lower than most other models as can be seen in Figure~\ref{fig-results}. 

\paragraph{Baselines.} The baselines for previous models trained on the Hateful Memes dataset are shown in Table~\ref{tab-results}.

\paragraph{Results.} Our top performing solutions are a simple average over probabilities using a single VL-BERT\textsubscript{LARGE} and one UNITER\textsubscript{LARGE+PA} (UNITER\textsubscript{LARGE} with the extra paired attention). We did not change the default training hyper-parameters of the vanilla pre-trained UNITER\textsubscript{LARGE} model, but only adapted the number of training steps according to the dataset size. Finally a Deep Ensemble of UNITER\textsubscript{LARGE+PA} models scored the highest performance. For this ensemble, we simply rerun the training with different random seeds and average over the predictions from each model. 
Table~\ref{tab-results} shows the top results for the final competition phase as well as the incremental improvements paired cross-attention brings to the UNITER model in the first phase.
The final results significantly outperform the baselines.

\begin{table}
\caption{Baselines from \citep{kiela2020hateful}. For our final models, we report the top performance scores, specifying both Accuracy and AUROC results.}
\label{tab-results}
\centering
\begin{tabular}{clcccc}
\toprule
 & & \multicolumn{2}{c} { Validation } & \multicolumn{2}{c} { Test } \\
Type & Model & Acc. & AUROC & Acc. & AUROC \\
\midrule
& Human & $-$ & $-$ & 84.70 & 82.65 \\
\midrule
& Image-Grid & 52.73 & 58.79 & 52.00 & 52.63 \\
Unimodal & Image-Region & 52.66 & 57.98 & 52.13 & 55.92 \\
& Text BERT & 58.26 & 64.65 & 59.20 & 65.08 \\
\midrule
& Late Fusion & 61.53 & 65.97 & 59.66 & 64.75 \\
& Concat BERT & 58.60 & 65.25 & 59.13 & 65.79 \\
& MMBT-Grid & 58.20 & 68.57 & 60.06 & 67.92 \\
Multimodal & MMBT-Region & 58.73 & 71.03 & 60.23 & 70.73 \\
(Unimodal Pretraining) & ViLBERT & 62.20 & 71.13 & 62.30 & 70.45 \\
& Visual BERT & 62.10 & 70.60 & 63.20 & 71.33 \\
\midrule
Multimodal & ViLBERT CC & 61.40 & 70.07 & 61.10& 70.03 \\
(Multimodal Pretraining) & Visual BERT COCO & 65.06 & 73.97 & 64.73 & 71.41 \\
\midrule
(Phase 1) & UNITER\textsubscript & $-$ & $-$ & \textbf{68.70} & \textbf{74.14} \\
(Phase 1) & UNITER\textsubscript{PA} & $-$ & $-$ & \textbf{68.30} & \textbf{75.29} \\
(Phase 1) & UNITER\textsubscript{PA} Ensemble & $-$ & $-$ & \textbf{66.60} & \textbf{76.81} \\
\midrule
(Phase 2) & VL-BERT + UNITER\textsubscript{PA} & \textbf{74.53} & \textbf{75.94} & \textbf{73.90} & \textbf{79.21} \\
(Phase 2) & UNITER\textsubscript{PA} Ensemble & \textbf{72.50} & \textbf{79.39} & \textbf{74.30} & \textbf{79.43} \\
\bottomrule
\end{tabular}
\end{table}

We note the most important findings:
\begin{itemize}
\item Single stream Transformer architectures pre-trained on the Conceptual Captions (CC) dataset \citep{sharma2018conceptual} show the best results and the deep ensemble further improves the overall performance. The choice of pre-training datasets matters in terms of domain similarity to the fine-tuning dataset, as also noticed by \citep{singh2020pretraining}.
\item We believe UNITER achieves higher results because it is pre-trained on the less noisy and superior annotations COCO dataset \citep{lin2015microsoft}. Similarly to the Hateful Memes dataset this is also high quality. As future work, we think it would be useful to investigate if pre-training VL-BERT on COCO would improve its results.
\item Interestingly, we find the paired attention approach only works for UNITER and not for any of the other models.
\item Training the large models from scratch performed poorly, as expected, due to the small dataset size.
\item We find the MMHS150K dataset is heavily biased towards hateful text and the keywords used to harvest it. The memes are not as subtle as the images in the Hateful Memes dataset. But perhaps it is a more vivid example of what lies out there in the meme wild.
\end{itemize}

\section{Conclusion}

In this work, we propose a simple yet effective set of techniques to help detect hate speech in a unique labeled dataset of high quality multimodal memes from Facebook AI. The goal is to identify hate speech using a multimodal model, while also being robust to the "benign confounders" that cause the binary label indicating whether a meme is hateful to flip.

We experiment with a number of large pre-trained Transformer based architectures and fine-tune both single stream state-of-the-art models such as VL-BERT, VLP and UNITER and dual stream models such as LXMERT. We compare their performance to the baselines provided by \citep{kiela2020hateful} and show all of the single-stream models significantly outperform these. We justify our choice for these transformers architectures by the possible advantages coming from the fact they were pre-trained on a wide spectrum of datasets from different domains. We also propose and adapt a novel bidirectional cross-attention mechanism to couple inferred caption information with the meme text obtained through optical character recognition. This addition achieves higher classification accuracy in labeling memes as hateful. Furthermore we show deep ensembles, a simple yet very powerful trick can improve on single model predictions by a significant margin.
As expected, we find training these large architectures from scratch performs poorly on a small set of examples such as the Hateful Memes dataset. However, we also find the choice of pre-training datasets also matters in terms of domain similarity to the fine-tuning dataset. 

We conclude that although the multimodal models are becoming increasingly sophisticated, there is still a large gap when comparing to human performance. This leaves considerable room for developing new algorithms to deal with multimodal understanding.

\medskip


\bibliographystyle{plainnat}
\bibliography{paper}

\end{document}